\documentclass[twoside]{article}

\makeatletter
\@namedef{ver@natbib.sty}{9999/12/31}
\let\setcitestyle\@gobble
\usepackage[accepted]{Styles/aistats2022}
\let\setcitestyle\undefined
\expandafter\let\csname ver@natbib.sty\endcsname\@undefined
\makeatother

\usepackage{multicol}
\usepackage{hyperref}
\usepackage{microtype}
\usepackage[center]{Styles/VABcommands}
\usepackage{Styles/VABenvironments}
\usepackage{Styles/VABcitations}
\usepackage{amsthm}
\usepackage{caption}
\usepackage{lipsum}
\usepackage{fancyvrb}
\usepackage{nicefrac}
\usepackage{subcaption}
\usepackage{comment}
\usepackage{array}
\usepackage{booktabs}
\usepackage{siunitx}
\usepackage{csquotes}
\usepackage{dblfloatfix}
\usepackage{float}
\MakeOuterQuote{"}

\newcommand{\distgeom}{{\sf d}}
\newcommand{\distalg}{{\sf d}_{\sf alg}}
\newcommand{\distk}[1]{{\sf d}_{#1}}
\newcommand{\disttwo}{\distk{2}}
\newcommand{\oscr}{{\sf o}}

\newcommand{\tableoneorder}[7]{ #1 & #4 & #2 & #5 & #3 & #6 & #7}
\newcommand{\tabletwoorder}[7]{ #1 & #3 & #5 & #4 & #2 & #6 & #7}

\bibliography{references.bib}

\runningtitle{Quadric Hypersurface Intersection for Manifold Learning in Feature Space}
\runningauthor{Pavutnitskiy, Ivanov, Abramov, Borovitskiy, Klochkov, Vialov, Zaikovskii, Petiushko}

\begin{document}

\twocolumn[

\aistatstitle{Quadric Hypersurface Intersection \\* for Manifold Learning in Feature Space}

\aistatsauthor{
\textbf{Fedor Pavutnitskiy}\textsuperscript{\ensuremath{*1}}
\\
\textbf{Artem Klochkov}\textsuperscript{\ensuremath{*2}}
\And
\textbf{Sergei O. Ivanov}\textsuperscript{\ensuremath{*2}}
\\
\textbf{Viktor Vialov}\textsuperscript{\ensuremath{*2}}
\And
\textbf{Evgeny Abramov}\textsuperscript{\ensuremath{*2}}
\\
\textbf{Anatolii Zaikovskii}\textsuperscript{\ensuremath{*2}}
\And
\textbf{Viacheslav Borovitskiy}\textsuperscript{\ensuremath{*2,4}}
\\
\textbf{Aleksandr Petiushko}\textsuperscript{\ensuremath{3}}
}
\aistatsauthor{}
\aistatsaddress{
\textsuperscript{\ensuremath{1}}HSE University
\quad
\textsuperscript{\ensuremath{2}}St. Petersburg State University
\quad
\textsuperscript{\ensuremath{3}}Lomonosov MSU
\\
\textsuperscript{\ensuremath{4}}St. Petersburg Department of Steklov Mathematical Institute of Russian Academy of Sciences}]

\begin{abstract}
    The knowledge that data lies close to a particular submanifold of the ambient Euclidean space may be useful in a number of ways.
    For instance, one may want to automatically mark any point far away from the submanifold as an outlier or to use the geometry to come up with a better distance metric.
    Manifold learning problems are often posed in a very high dimension, e.g. for spaces of images or spaces of words.
    Today, with deep representation learning on the rise in areas such as computer vision and natural language processing, many problems of this kind may be transformed into problems of moderately high dimension, typically of the order of hundreds.
    Motivated by this, we propose a manifold learning technique suitable for moderately high dimension and large datasets.
    The manifold is learned from the training data in the form of an intersection of quadric hypersurfaces---simple but expressive objects.
    At test time, this manifold can be used to introduce a computationally efficient outlier score for arbitrary new data points and to improve a given similarity metric by incorporating the learned geometric structure into it.
\end{abstract}

\section{INTRODUCTION}

One particularly interesting new area of research for manifold learning is motivated by the recent advances in deep representation learning. In a wide range of industrial scenarios where deep feature extractor is used as a part of a larger pipeline, a feature space level outlier detector may help tackle the problem of out-of-distribution input data at test time, which, in its turn, may appear due to undertraining, faulty preprocessing or even a deliberate attack.
Manifold learning may be used to build such a detector.
Moreover, in problems where we need to compare the similarity of different inputs, e.g. in face recognition, geometry-based detector can be used to improve the similarity metric.

Motivated by these problems, we propose a manifold learning technique where the manifold is learned in form of an intersection of quadric hypersurfaces---the zero-sets of quadratic polynomials. 
Like principal component analysis (PCA), it yields a manifold as a subset of the ambient~Euclidean~space.

\begin{table}[b!]
\vspace*{-3.575ex}
\footnotesize\urlstyle{same}\textsuperscript{\ensuremath{*}}Equal contribution. Code available at: \textsc{\url{http://github.com/spbu-math-cs/Quadric-Intersection}}. \\* Correspondence to: \href{mailto:fpavutnitskiy@hse.ru}{\textsc{fpavutnitskiy@hse.ru}}.
\vspace*{3.425ex}
\end{table}

Fitting a quadric hypersurface intersection is posed as an optimization problem of minimizing distances from training dataset to the intersection.
Since the geometric distances are computationally expensive to calculate we discuss various approximations.
The simplest possible choice gives rise to a close relative of the kernel PCA with quadratic kernel.
We are going beyond this basic model and utilize a finer and more robust approximation.
Moreover, we introduce new optimization constrains to make the optimization problem equivariant with respect to isometric transformations of the training dataset.

The proposed quadric hypersurface intersection model is much more expressive than the linear one used in PCA, which is the intersection of hyperplanes.
It is also more robust, geometry-respecting and scalable than simple variations of kernel PCA.
The number of parameters that define the quadric intersection model grows quadratically with the dimension, thus making it suitable for moderately high-dimensional spaces, e.g. for feature spaces of deep models.
One of the most important features of the proposed technique is that it is amenable to stochastic gradient descent~(SGD), which allows (sub)linear scaling with respect to the training dataset size and is straightforward to implement using modern automatic differentiation frameworks.\footnote{An additional justification for the expressiveness of the model can be found in \textcite[Theorem 1]{mumford2010varieties}.}

To showcase the potential of the proposed technique, in Section~\ref{sec:experiments} we
consider its application to an industrial level image classification and outlier detection problem.

\section{SETTING AND RELATED WORK}\label{sec:target_setting}

We aim to propose a manifold learning technique to drive a geometry-based outlier detector which may be used at a feature space level of industrial scale deep representation learners.
We are looking at large unlabeled datasets of synthetically structured data and of moderately high dimension~(order of hundreds).
We~need to keep in mind that training data may be contaminated and may possess complex topology, for instance be highly clustered, with unknown or even non-fixed number of clusters (this is common in open set classification problems).
Thus, the technique should be expressive, robust and scalable.

The need for a new technique comes from the fact that classical manifold learning algorithms, such as Isomap \cite{tenenbaum2000global}, LLE \cite{roweis2000nonlinear}, Laplacian eigenmaps \cite{belkin2001laplacian}, LTSA \cite{zhang2004principal} and anomaly detection techniques based on them (e.g. \textcite{hein2006manifold}) are not readily suitable for large-scale problems without further modifications.

Geometry-motivated anomaly detection techniques like one-class SVM~(OCSVM, \textcite{scholkopf1999support}), support vector data description~(SVDD, \textcite{tax2004support})\footnote{In most cases this approach is actually equivalent to OCSVM~\cite{lampert2009kernel}.} or the kernel PCA based novelity detector~\cite{hoffmann2007kernel} aim to solve similar problems.
However, they usually fail to be simultaneously expressive, robust and, most importantly, scalable enough.
To prove this point, we evaluate our technique against~(the suitable approximations of) these methods in Section~\ref{sec:experiments}.

We acknowledge that the idea of describing a point cloud as a zero set of polynomial functions is not novel.
As will be explained later, even the simple PCA may be interpreted this way.
More recently, \textcite{livni2013vanishing} used similar ideas for solving classification problems.
However, their singular value decomposition based approach is not scalable enough for our target setting and thus not really relevant to our further developments.
We also acknowledge the related recent papers by~\textcite{li2017efficient} and~\textcite{jung2012analysis}.

\section{MANIFOLD LEARNING}
\emph{Manifold learning}, as a term, refers to a diverse collection of techniques motivated by the \emph{manifold hypothesis}~\cite{fefferman2016testing}, the statement that natural datasets (e.g. images of pets) lie in the vicinity of a relatively low-dimensional manifold embedded in a higher-dimensional ambient space.
Manifold learning is often considered synonymous to \emph{nonlinear dimensionality reduction} \cite{lee2007nonlinear}, though the latter more often refers to data-visualization methods.

There exists a large set of manifold learning techniques, many of them are considered by \textcite{ma2011manifold}.
Most of these techniques can be thought of as black boxes which take a point cloud in a high-dimensional Euclidean space and which map every point of the cloud into a point in a low-dimensional Euclidean space.
For example, multidimensional scaling algorithms seek the mapping so as to preserve pairwise distances as well as possible, while PCA, viewed through an appropriate lens, tries to preserve most of the data's variation.

Another natural but less-often studied class of manifold learning techniques tries to characterize the manifold in the vicinity of which the point cloud lies as a submanifold of the ambient Euclidean space.
We highlight that this shift in formulation allows one to ask additional questions, such as how far an arbitrary point in the ambient space is from the manifold---a key question for the outlier detection applications.
We proceed to discuss this formulation further.

\subsection{\emph{Characterizing} Manifolds}

We begin by recalling and highlighting a key property of principal component analysis, namely that it characterizes the manifold it finds as a submanifold of the ambient Euclidean space.
Given a centered point cloud $\v{p}_1, \dots, \v{p}_n \in \R^d$, PCA finds orthonormal vectors $\v{v}_1, \dots, \v{v}_d \in \R^d$ such that
\[ \label{eqn:pca_space}
V_k = \Set*{\v{x} \in \R^d}{\v{x} = \alpha_1 \v{v}_1 + \dots + \alpha_k \v{v}_k}
\]
is the $k$-dimensional linear subspace (thus a submanifold) of the ambient Euclidean space $\R^d$ that optimally fits the point cloud in a suitable sense.
With this definition, it is possible to compute the distance from any point $\v{p} \in \R^d$ to the closest point of $V_k$:
\[ \label{eqn:pca_distance}
\distgeom(\v{p}, V_k) = \norm[2]{\v{p} \!-\! \sum_{j=1}^k \innerprod{\v{p}}{\v{v}_j} \v{v}_j} = \del[1]{\!\!\sum_{j=k+1}^d\!\! \innerprod{\v{p}}{\v{v}_j}^2}^{1/2},
\]
where $\norm{\cdot}$ and $\innerprod{\cdot}{\cdot}$ are standard Euclidean norm and inner product, respectively.
In this sense, PCA explicitly characterizes the manifold through~\eqref{eqn:pca_space}.
Hereinafter we use $\distgeom$ to denote the \emph{geometric distance}: for a subset $X\subset \R^d$ and a point $\v{p} \in \R^d$ this distance is given by
\[
\distgeom(\v{p},X) = \inf_{\v{x} \in X} \norm{\v{p} - \v{x}}.
\]

PCA's way of characterizing a manifold is very convenient but relies on the fact that elements of a linear subspace can be represented as linear combinations of a finite collection of basis vectors, which does not directly extend to non-linear domains.
However, one can modify this point of view, to make it more amenable to the non-linear setting by considering the linear subspace that PCA finds as the zero set of some vector-valued linear mapping. For a map $\v{F}:\R^d\to \R^l$ we set 
\[
Z(\v{F})=\Set*{\v{x} \in \R^d}{\v{F}(\v{x})=0}.
\]
Then we have $V_k = Z(\v{F}^{(k)})$ for $\v{F}^{(k)}: \R^d \to \R^{d-k}$ given by $\v{F}^{(k)} (\v{x}) \!=\! \del{\innerprod{\v{x}}{\v{v}_{k+1}}, \ldots, \innerprod{\v{x}}{\v{v}_{d}}}$. 
This is an instance of a very general way of representing a submanifold, as a zero set of some smooth function: rather than viewing a manifold as the span of a set of basis vectors, we can view a manifold as a solution to the system of equations $Z(\v{F}) = 0$ for a suitable $\v{F}$.

By the preimage theorem \cite[\S 2, Lemma 1]{milnor1997topology}, under mild technical assumptions on $\v{F}$, we have that $Z(\v{F})$ is indeed a manifold.

In the following, we will rely on representing a manifold as a zero set of a function, thus shifting the problem of finding a manifold to the problem of finding a function.

Note that this representation is much more expressive (though less explicit) than representing a manifold as an image of $\R^k$ under some function $\v{G}: \R^k \to \R^d$ as for the mapping $\v{G}(\alpha_1, .., \alpha_k) = \alpha_1 \v{v}_1 + \dots + \alpha_k \v{v}_k$ in PCA.
Although this method is widely used, for example in autoencoder neural networks, it can only represent manifolds that can be covered by a single chart: this prevents the accurate representation of, for instance, disconnected domains or of the simple sphere or torus.

\section{INTERSECTIONS OF QUADRICS}

As noted in the previous section, the $k$-dimensional submanifold of $\R^d$ that PCA finds from a point cloud $\v{p}_1, .., 
\v{p}_n \in \R^d$ is a zero set $Z(\v{F}^{(k)})$ of some linear function $\v{F}^{(k)}: \R^d \to \R^{d-k}$.
If we look a little bit deeper at how this manifold is defined, we can see that $\v{F}^{(k)}$ solves the optimization problem
\[ \label{eqn:pca_optimization}
\v{F}^{(k)} = \argmin {\sum}_{j=1}^n \distgeom (\v{p}_j, Z(\v{F}^{(k)}))^2.
\]
Set $\v{u}_j=\v{v}_{j+k}$ for indices $1\leq j\leq d-k.$
Then ${\v{F}^{(k)} = (f_1, .., f_{d-k})}$ where $f_j(\v{x}) = \innerprod{\v{x}}{\v{u}_{j}}$ are linear polynomials with coefficients $\v{u}_{j} \in \R^d$ such that vectors~$\v{u}_1, \dots, \v{u}_{d-k}$ are orthonormal.
It means that~\eqref{eqn:pca_optimization} has to be optimized over $d-k$ vectors $\v{u}_j$ of dimension~$d$ under the constraint that they should form an orthonormal system.
Expanding the distance in~\eqref{eqn:pca_optimization} through~\eqref{eqn:pca_distance}, we get a simple optimization problem which can either be solved exactly by computing the singular value decomposition (SVD) of a $d \x n$ matrix, or approximately through gradient-based optimization.

We propose to extend this by considering quadratic polynomials $f_j$ instead of the linear ones, as components of function $\v{F} = (f_1,\dots , f_{d-k})$.

The zero set of a quadratic polynomial (polynomial of degree~$2$) is called a \emph{quadric hypersurface} or simply a \emph{quadric}.
We consider linear polynomials and constants to be special cases of quadratic polynomials and similarly refer to their zero sets as quadrics.
The word \emph{hypersurface} is justified by the fact that in the non-degenerate case $Z(f_j)$ are \mbox{$(d-1)$-dimensional}.\footnote{
A quadric is non-degenerate if the Hessian matrix of the homogenization $f^{\sf hom}(x_1,\dots,x_{d+1})=x_{d+1}^2f(x_1/x_{d+1},\dots,x_d/x_{d+1})$ of the corresponding polynomial~$f$ is non-singular.
In this case the quadric is a smooth algebraic variety and thus a manifold of dimension~${d-1}$ \cite[Example 3.3]{harris2013algebraic}.
}

Moreover, the zero set~$Z(\v{F})$, which coincides with the intersection $Z(\v{F}) = Z(f_1) \cap \dots \cap Z(f_{d-k})$ is usually a $k$-dimensional manifold.
Intuitively, each of the $d-k$ quadrics eliminates one dimension of the $d$-dimensional ambient space.
We do not dwell here on a precise condition for intersection to be a $k$-dimensional manifold.

\begin{figure*}[t]
\centering
\includegraphics[width=0.28\linewidth]{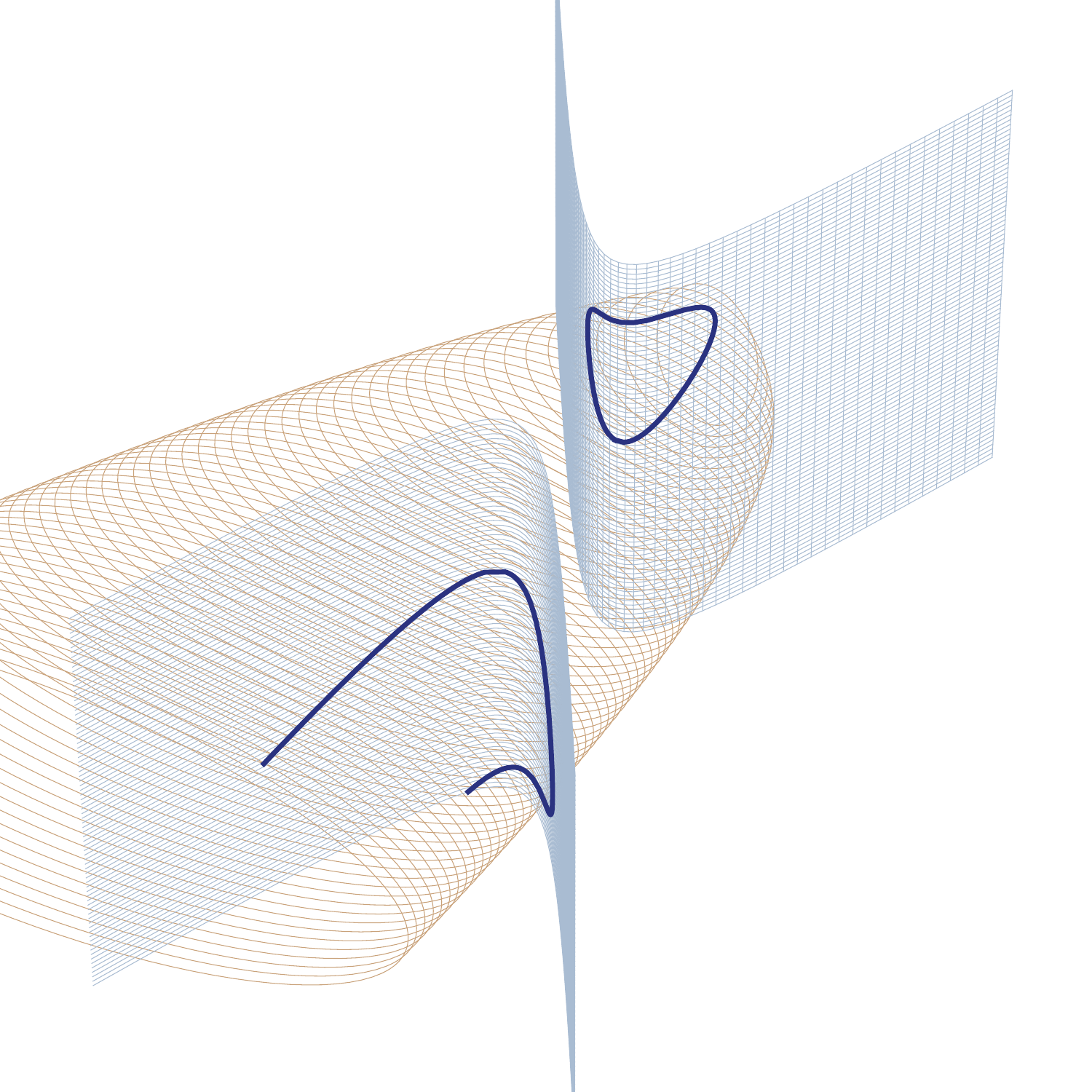}%
\hspace{0.07\textwidth}%
\includegraphics[width=0.28\linewidth]{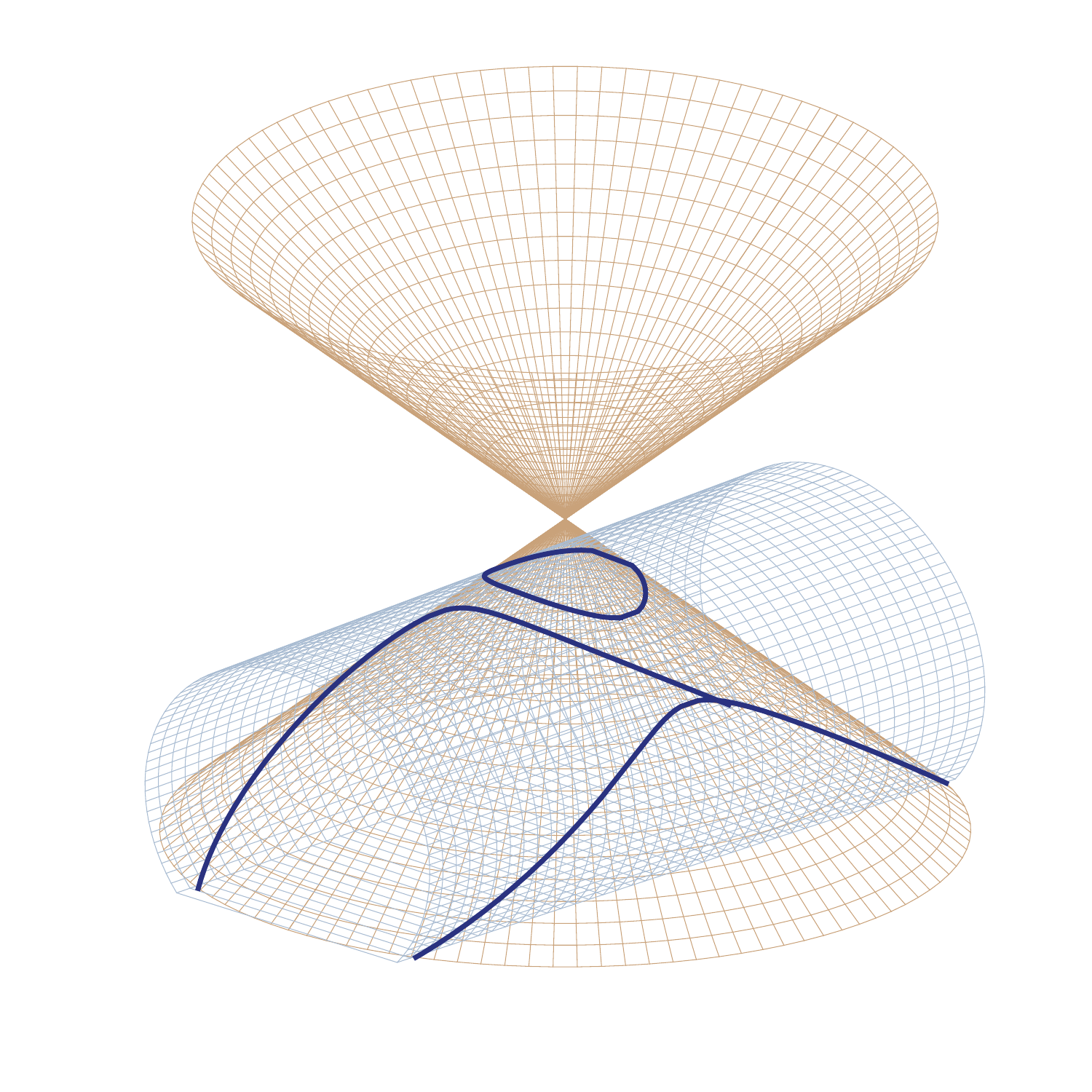}%
\hspace{0.07\textwidth}%
\includegraphics[width=0.28\linewidth]{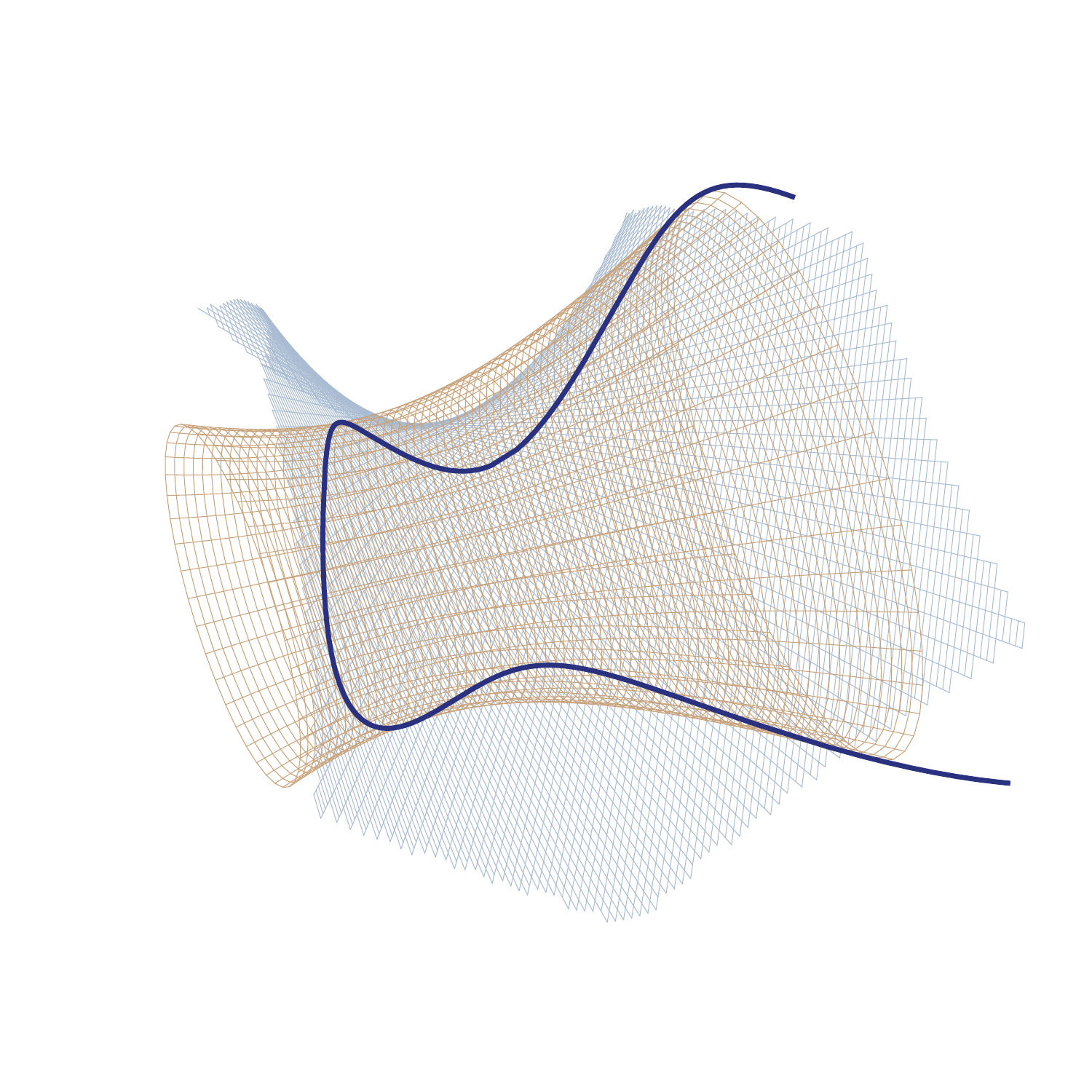}%
\caption{Examples of quadrics and their intersections in $\R^3$. Two quadric surfaces are depicted by two meshes, each with its own color. The dark blue line portrays their intersection. From left to right: the intersection of an elliptic paraboloid with a hyperbolic cylinder; the intersection of an elliptic cone with an elliptic cylinder; the intersection of a hyperbolic hyperboloid and a hyperbolic paraboloid.}
\label{fig:ex}
\end{figure*}

Quadrics in $\R^2$ are conic sections: ellipses, hyperbolas and parabolas.
However already in $\R^3$ quadrics and their intersections may be much less trivial.
This is illustrated on Figure~\ref{fig:ex}.
We refer the reader to the paper of \textcite{beale2016fitting} and to the references therein for a brief review of the previous work on the topic of fitting quadrics in low dimension, while proceeding to present an algorithm suitable for high dimension and large datasets.

\subsection{Fitting a Quadric Intersection} \label{sec:the_basic_technique}

We start with the optimization problem~\eqref{eqn:pca_optimization}, which, at least in theory, is as applicable for intersections of quadrics as it is to linear subspaces (intersections of linear hypersurfaces).
Any quadratic polynomial in $\R^d$ can be written as
\[\label{eqn:quadractic_polynomial}
f(x_1,\dots,x_d)={\sum}_{i\leq j} \alpha_{i,j}x_ix_j+{\sum}_{i} \alpha'_i x_i + \alpha'',
\]
where $\alpha_{i,j},\alpha'_i,\alpha''\in \R$.
We denote the vector of its coefficients by ${\v{v}(f)=(\alpha_{1,1}, ..,\alpha_{d,d}, \alpha'_1, .., \alpha'_d,\alpha'')^\top \in \R^D}$, where $D = \del{ d^2 + 3 d} / 2 + 1$ is the number of monomials of degree $\leq 2$ in $d$ variables.
Thus the intersection of quadrics $Z(f_1), \dots, Z(f_m)$ is determined by $m$ vectors $\v{v}(f_1), \dots, \v{v}(f_m)$, similarly to how a linear subspace is determined by the basis vectors in PCA.

The intersection of quadrics $Z(\v{F})$ does not change if we make a replacement $f_i \mapsto a f_i$ for $a \ne 0$ or $f_i \mapsto f_i + a f_j$ for any $a \in \R$ and $i \ne j$, which is quite clear from viewing the $Z(\v{F})$ as the solution set of a system of equations.
Moreover, the problem~\eqref{eqn:pca_optimization} per se has $\v{F}(x) \equiv 0$ for a trivial solution that corresponds to the whole $Z(\v{F}) = \R^d$.
To handle both of these issues we need to introduce~some constraints.
We~therefore look for an {\it orthonormal} collection of quadratic polynomials with respect to some inner product, the most simple of which is the standard inner product of vectors $\v{v}(f_1), ..,\v{v}(f_m)$.

It turns out though that optimization problem~\eqref{eqn:pca_optimization} is very hard to solve for quadrics: computing even a single $\distgeom(\v{p}_j, Z(f_k))$ is a (computationally) hard problem because projecting a point onto a quadric is nontrivial.
A number of approximations have therefore been proposed \cite{taubin1991estimation, taubin1993improved} such as ${\distalg(\v{p}_j, Z(f_k)) := \abs{f_k(\v{p}_j)}}$, called the \emph{algebraic distance}.\footnote{A number of techniques in the literature may be linked to the algebraic distance, e.g.~\textcite{coope1993circle}.}
Approximating ${\distgeom(\v{p}_j, Z(\v{F})) \approx \del[1]{\sum_{k=1}^m \abs{f_k(\v{p}_j)}^2}^{1/2}}$, gives the optimization problem ($\delta_{k l}$ is the Kronecker delta):
\[ \label{eqn:alg_dist_optimization}
(f_1, \dots, f_m) = \argmin_{\substack{\innerprod{\v{v}(f_k)}{\v{v}(f_l)} = \delta_{k l} \\ \text{for } k, l = 1, \dots, m}}
\sum_{j=1}^n \sum_{k=1}^m \abs{f_k(\v{p}_j)}^2
.
\]
In Appendix \ref{appdx:connection_to_PCA} we show that the problem \eqref{eqn:alg_dist_optimization} is deeply related to the polynomial kernel PCA.
Despite this connection, we view this problem from a different angle and propose a simple yet fruitful idea of solving~\eqref{eqn:alg_dist_optimization} by applying stochastic gradient descent to perform \emph{unconstrained} optimization of the corresponding Lagrangian, treating the Lagrange multiplier as a tunable hyperparameter.
Building on this idea, we proceed to improve this approach. But first, we discuss its downsides.

\paragraph{Discussion} 
The problem \eqref{eqn:alg_dist_optimization} has a number of downsides.
First, the algebraic distance is a poor approximation of the geometric distance.
In practice, this may result in artifacts and unstable behavior of gradient-based optimization.
Second, the technique's deep connection to PCA suggests that it may be very sensitive to outliers, similar to how PCA is \cite{candes2011robust}.
Finally, the optimization problem \eqref{eqn:alg_dist_optimization} does not reflect well the geometric structure of the manifold learning problem as it is non-equivariant in the following sense.
If~$f_1,\dots,f_m$ is an orthonormal collection of quadratic polynomials and $\theta:\R^d\to \R^d$ is an isometry, then the collection~$f_1\circ \theta,\dots,f_m \circ \theta$ can fail to solve the optimization problem~\eqref{eqn:alg_dist_optimization} for the point cloud transformed by $\theta$ (see Appendix \ref{appdx:non-equivariance}).
Below we address these downsides and present a new technique for fitting an intersection of quadrics to~a~point~cloud.

\subsection{Loss Function} \label{sec:improved_loss}
In \textcite{taubin1993improved}, a notion $\distk{k}(\v{p}, Z(f))$ of \emph{approximation distance of order $k$} is defined, building upon the idea of $k$-th order Taylor approximation.
This distance coincides with the only non-negative root of a certain polynomial $c_0+c_1t+\dots+c_kt^k$ of degree~$k$, whose coefficients depend on partial derivatives of $f$ in $\v{p}$---full details are given in Appendix~\ref{appendix_distances}.
One particularly popular approximation of a distance is the \emph{distance of order 1} given by $\distk{1}(\v{p}, Z(f)) = \abs{f(\v{p})}/\norm{\grad f (\v{p})}$.
For a quadratic polynomial $f$, the $k$-distance coincides with the $2$-distance for $k\geq 2$.
Moreover, there is a simple explicit formula for the $2$-distance:
\[
\disttwo(\v{p},Z(f))
=
\del[1]{\sqrt{h^2 + \abs{f(\v{p})} \norm{f}_{HS}} - h}/\norm{f}_{HS}
,
\]
where $h = \norm{\grad f(\v{p})}/2$, and $\norm{f}_{HS}$ is a certain Hilbert--Schmidt norm defined in \eqref{eqn:hs_definition} below.

The distance of order $2$ gives a better approximation of the geometric distance in a number of ways.
Firstly, it is, in contrast to the algebraic distance, equivariant.
More precisely, for isometries $\theta:\R^d\to \R^d$ we have
\[
\disttwo(\theta(\v{p}),Z(f \circ \theta))=\disttwo(\v{p},Z(f))
.
\]
The proof of this fact can be found in Appendix \ref{appdx:equivariance}.
Secondly, it is majorized by the geometric distance
\begin{equation}\label{eq:d2_ineq}
\disttwo(\v{p},Z(f)) \leq \distgeom(\v{p},Z(f))
,
\end{equation}
(see \textcite[\S 6]{taubin1991estimation}) while $\distalg$ and $\distk{1}$ are not, the latter may even be infinite.
This limits the contribution of outliers to the optimization objective and thus facilitates robustness. Indeed, Equation~\eqref{eq:d2_ineq} shows that the $2$-distance cannot  be arbitrarily large for points that are not geometrically far away from the manifold. This is an advantage over $1$-distance: at every point where the gradient $\nabla f$ vanishes, $1$-distance from the quadric $Z(f)$ to this point will be infinite.  
Thirdly and lastly, this distance is simple to compute.
Hence, for quadrics, the distance of order $2$ constitutes the optimal candidate approximation of the geometric distance.

Recall that in order to define a new optimization objective, we have to approximate the distance $\distgeom(\v{p}, Z(\v{F}))$ for $\v{F} = (f_1, .., f_m)$, not just the distance $\distgeom(\v{p}, Z(f_k))$.
In the original optimization problem \eqref{eqn:alg_dist_optimization}, we used the $l^2$-based term $\del[0]{\sum_{k=1}^m \abs{\distalg(\v{p}, Z(f_k))}^2}^{1/2}$ as a proxy for $\distgeom(\v{p}, Z(\v{F}))$.
Here, we suggest to use the $l^1$-based term $\sum_{k=1}^m \abs{\disttwo(\v{p}, Z(f_k))}$, based on the consideration that $l^1$-loss is more robust to outliers than the squared $l^2$-loss.
This change also leads to equivariance of the objective, as equivariance of each term of the sum implies equivariance of the whole sum.

\subsection{Constraints} \label{sec:improved_constraints}

By replacing $\distalg$ with $\disttwo$, we have made the optimization objective equivariant with respect to the action of the Euclidean group.
Unfortunately, the coefficient-wise inner product for quadratic polynomials is not equivariant and the constrained optimization problem still exhibits geometrically unnatural behavior.
To resolve this, we suggest an inner product for quadrics such that the constraint of orthonormality with respect to it makes the whole optimization problem equivariant.

Any quadratic polynomial $f$ can be represented in form
\[ \label{eqn:polynomial_matrix_form}
f(\v{x}) = \v{x}^\top \m{A} \v{x} + {\bf b}
\v{x} + {\bf c},
\]
where $\m{A}$ is a symmetric $d\times d$ matrix, $b$ is a row vector, $c \in \R$.
If ${f(x_1, ..,x_d) = \sum_{i \leq j} \alpha_{i,j} x_i x_j + \sum_{i} \alpha'_i x_i + \alpha''}$, as in \eqref{eqn:quadractic_polynomial} then $\m{A}_{i,i} \!=\! \alpha_{i,i}$ and $\m{A}_{i,j} \!=\! \m{A}_{j,i} \!=\! \alpha_{i,j}/2$, $i \!<\! j$.

If $f$ and $g$ are quadratic polynomials with corresponding symmetric matrices $\m{A}$ and $\m{B}$, we define their Hilbert--Schmidt (degenerate) inner product as the Hilbert--Schmidt inner product of their matrices
\[ \label{eqn:hs_definition}
&\innerprod{f}{g}_{HS}
\!=\!\!
{\sum}_{i,j}\! \m{A}_{ij} \m{B}_{ij}
,
&
& \norm{f}_{HS} \!=\!\! \sqrt{\innerprod{f}{f}_{HS}}.
\]
This inner product is degenerate in the sense that the corresponding norm $\|f\|_{HS}$ is actually only a seminorm, i.e. $\|f\|_{HS} = 0$ does not imply $f = 0$, this is because it vanishes on polynomials of degree $\leq 1$.
A collection of quadratic polynomials $f_1,\dots,f_m$ is called HS-orthonormal if it is orthonormal with respect to the Hilbert--Schmidt inner product.
In particular, an HS-orthonormal collection of quadratic polynomials consists only of polynomials of degree $2$.\footnote{Note that such a collection can still be used to represent a linear subspace, simply because a linear equation of form ${f(x) = 0}$ may be transformed into the quadratic equation ${f(x)^2 = 0}$ with the same solution.}

It is easy to check (see Appendix~\ref{appdx:equivariance_HS} for details) that this inner product is equivariant with respect to the action of the Euclidean group.
Specifically, for any isometry ${\theta: \R^d\to \R^d}$ the following holds:
\[
\innerprod{f}{g}_{HS}
=
\innerprod{f \circ \theta}{g \circ \theta }_{HS}
.
\]

If we define the weighted vector of coefficients by ${\tilde{\v{v}}(f) = (\alpha_{1,1}, \alpha_{1,2}/\sqrt{2}, \dots, \alpha_{d-1,d}/\sqrt{2}, \alpha_{d,d})^\top}$,
where all coefficients that correspond to the non-diagonal entries of $A$ are divided by $\sqrt{2}$, then, we have ${\innerprod{f}{g}_{HS} = \innerprod{\tilde{\v{v}}(f)}{\tilde{\v{v}}(g)}}$, with the regular Euclidean inner product on the right-hand side.

There is a number of ways to enforce orthonormality of $\tilde{\v{v}}(f_1),\dots,\tilde{\v{v}}(f_m)$.
The problem is well-studied in the context of orthogonality of filters inside layers of neural networks \cite{Bansal2018regularizations}.
We propose to use the soft orthogonality regularization term ${\norm[0]{\tilde{V}(\v{F})^T \tilde{V}(\v{F}) - I}^2_{HS}}$, where~${\tilde{V}(\v{F}) =(\tilde{\v{v}}(f_1),\dots,\tilde{\v{v}}(f_m))}$ is a $\nicefrac{d(d+1)}{2} \times m $-matrix, whose columns are $\tilde{\v{v}}(f_i)$.

\subsection{Summary: an Outlier-robust and Equivariant Algorithm}
\label{sec:improved_technique}

Assume that we have a cloud of points ${\v{p}_1,\dots, \v{p}_n\in \R^d}$ and we want to find $m$ quadratic polynomials $f_1,\dots,f_m$ so that this cloud lies close to the intersection of quadrics $Z(f_1)\cap \dots \cap Z(f_m)$.
Recalling that~$\delta_{k l}$ denotes the Kronecker delta, we formally pose the optimization problem as follows
\begin{equation}\label{eq:our_optimization_problem}
   \del{f_1, .., f_m} =\!\!\!\! \argmin_{\substack{\innerprod{f_k}{f_l}_{HS} \,=\, \delta_{k l} \\ \text{for } k, l = 1, \dots, m}} 
   \sum_{j=1}^n \sum_{k=1}^m \abs{\disttwo(\v{p}_j, Z(f_k))}, 
\end{equation}

where each of $m$ quadrics $f_k$ is represented by the $D$-dimensional vector of its coefficients.
Since both the optimization objective and the constraints are equivariant, the whole optimization problem is equivariant: if quadrics $f_1, \dots, f_m$ solve the optimization problem above, then for any isometry $\theta:\R^d\to \R^d$ the collection $f_1 \circ \theta, \dots, f_m \circ \theta$ solves the same optimization problem for the transformed point cloud $\theta(\v{p}_1),\dots, \theta(\v{p}_n)$.

Compared to the optimization problem \eqref{eqn:alg_dist_optimization}, closely related to the polynomial kernel PCA, problem~\eqref{eq:our_optimization_problem} facilitates robustness and is equivariant.\footnote{Non-equivariance of the original kernel PCA optimization problem \eqref{eqn:alg_dist_optimization} is shown in Appendix \ref{appdx:non-equivariance}.} This is due to a finer distance approximation and $l^1$-averaging in the loss and due to Hilbert--Schmidt equviariant constraints.
The latter is shown theoretically in Appendix~\ref{appdx:equivariance_HS} and the former is illustrated by a toy example in Section~\ref{sec:toy_experiment}.

In practice, we employ a soft constraint incorporated into the loss that is given by the Lagrangian\footnote{For an additional discussion on the choice of the regularization see Appendix~\ref{appdx:ir_hyperparameters}.}
\begin{equation}\label{eq:concrete_loss}
\sum_{j=1}^n \sum_{k=1}^m \abs{\disttwo(\v{p}_j, Z(f_k))}
+
\lambda
\norm{\tilde V(\v{F})^T \tilde V(\v{F}) \!-\! I}^2_{HS},
\end{equation}
where $\tilde V(\v{F})$ is as above and $\lambda$ is a hyperparameter.
We solve this via the stochastic gradient descent over the $D\!\x\!m$-dimensional set of quadric coefficients.
Thanks to this, we have (sub)linear scaling with respect to data size and ease of implementation---another two key features of the approach.

\subsection{Out-of-distribution Detection}\label{subsec:ood}

Assuming a moderately high (order of hundreds) dimensional feature space, we may fit an intersection of quadrics to the feature embeddings of the training data, as described in the previous section.
The assumption that the embeddings lie close to the found manifold suggests the distance to manifold as a natural~outlier~score.

Since it is computationally difficult to evaluate this distance exactly, we suggest using the same approximation that was utilized for training.
Specifically, we define the outlier score $\oscr(\v{p})$ of an arbitrary point $\v{p}$ in the embedding space~by
\[ \label{eqn:outlier_score}
\oscr(\v{p}) = \frac{1}{m} \sum_{k=1}^m \disttwo (\v{p}, Z(f_k)),
\]
where $f_1, \dots, f_m$ are the quadratic polynomials that define the found manifold.
This average, while easy to compute, may serve as an effective out-of-distribution score.
In Section~\ref{sec:exp_ood_and_robust} we evaluate the performance of the out-of-distribution detector built upon it.\footnote{Note that this score may alternatively be viewed as the combined score of the ensemble of simple out-of-distribution detectors induced by individual quadrics.}

\subsection{Similarity Robustification}\label{subsec:similarity_robust}

The most natural way to incorporate the geometric structure of a manifold into the similarity measurement procedure is to use the geodesic distance of the manifold as the new dissimilarity function.
Unfortunately though, computing the geodesic distance between a pair of points on the intersection of quadrics is a difficult problem rendering such an approach impractical.

On the other hand, a different approach can be used to improve a given (dis)similarity metric (e.g. Euclidean) using the found geometric structure.
Namely, by incorporating the information of the outlierness into the similarity function, we can make classification more robust.
For instance, we can modify a similarity function~$s$ by declaring the outliers dissimilar to anything:
\[ \label{eqn:robustification_truncation}
s_h(\v{x},\v{y}) =
\begin{cases}
s(\v{x},\v{y}), &\max(\oscr(\v{x}), \oscr(\v{y})) < t, \\
0, &\max(\oscr(\v{x}), \oscr(\v{y})) \geq t,
\end{cases}
\]
where $t$ is a threshold hyperparameter balancing precision and recall and $s_h$ is the new robustified similarity.
This approach is evaluated in Section~\ref{sec:exp_ood_and_robust} along with the out-of-distribution detector described above.

Apart from the empirical results from Section~\ref{sec:experiments}, the robustification given by~\eqref{eqn:robustification_truncation} is supported by the observation that the similarity between outliers is often abnormally large.
This may be seen as a consequence of deep feature extractors’ inability to distinguish the out-of-distribution samples.

\section{EXPERIMENTS} \label{sec:experiments}

\subsection{Toy Example} \label{sec:toy_experiment}

\begin{figure*}[t]
\centering
    \begin{subfigure}{0.32\textwidth}
        \includegraphics[trim={0.5cm 3cm 0.5cm 2cm},clip, width=\textwidth]{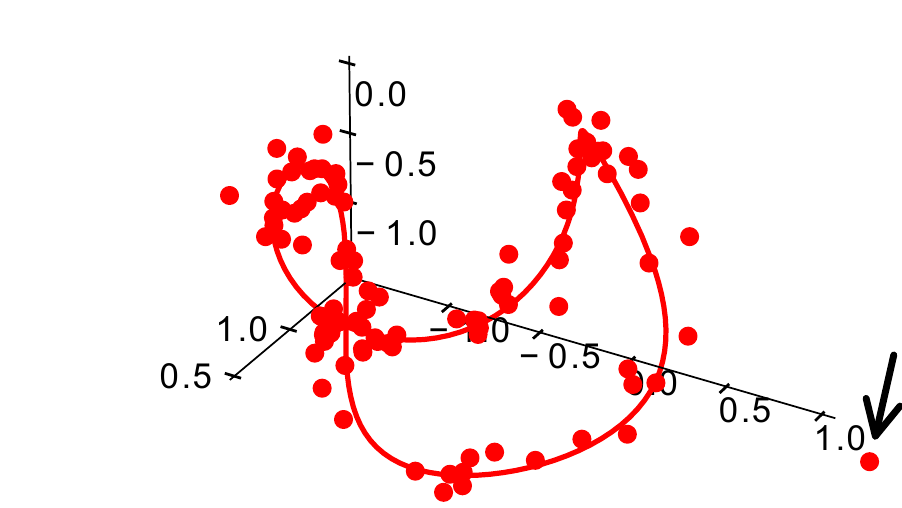}
        \caption{The ground truth}
    \end{subfigure}
    \begin{subfigure}{0.32\textwidth}
        \includegraphics[trim={0.5cm 3cm 0.5cm 2cm},clip,width=\textwidth]{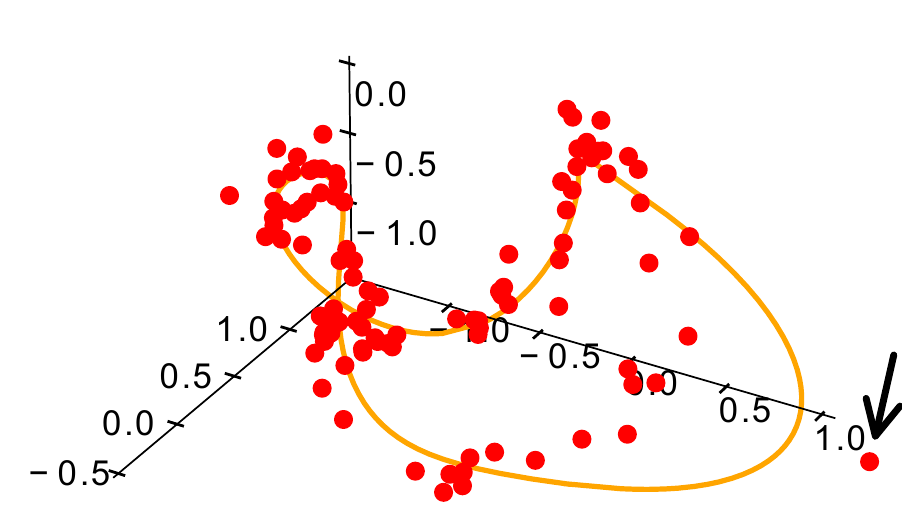}
        \caption{Kernel PCA fit}
    \end{subfigure}
    \begin{subfigure}{0.32\textwidth}
        \includegraphics[trim={0.5cm 3cm 0.5cm 2cm},clip,width=\textwidth]{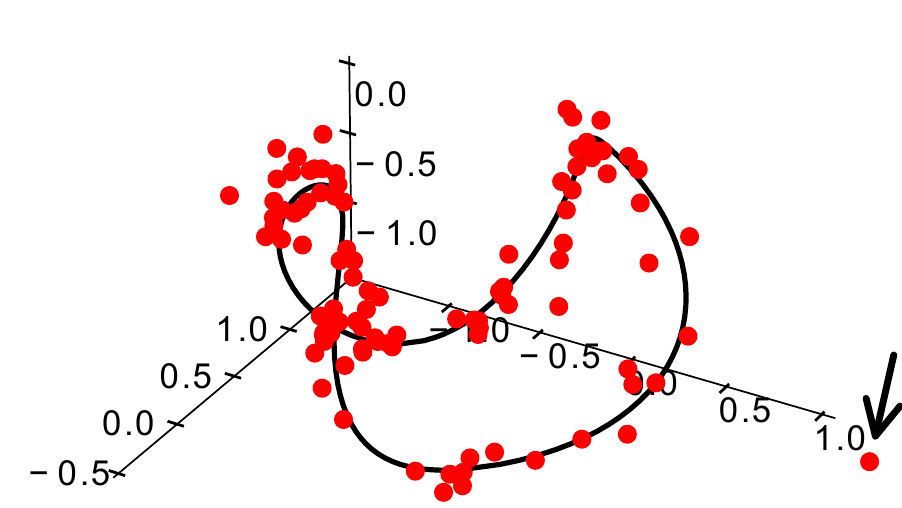}
        \caption{The proposed technique fit}
    \end{subfigure}
\caption{
The point cloud of $99$ points is generated by adding Gaussian noise to random points lying on the ground truth curve.
A random outlier (under the arrow) having twice the norm of the points on the curve is then added to the point cloud.
The ground truth (a), the fit corresponding to the exact solution of the optimization problem \eqref{eqn:alg_dist_optimization} from Section~\ref{sec:the_basic_technique} computed by means of kernel PCA with quadratic kernel (b) and the fit of the proposed technique summarized in Section \ref{sec:improved_technique} (c) are shown.
}
\label{fig:toy_example}
\end{figure*}

To illustrate the robustness of the proposed technique we study a toy example.
Consider the curve shaped like a seam line of a tennis ball, given by the parametrization
\[
&x = a\cos t+ b\cos 3t,
&
&y = a\sin t - b \sin 3t,
\\
&z = 2\sqrt{ab} \sin 2t,
&
&\text{with } a = 0.8, b = 0.2,
\]
and take a cloud of $100$ points generated by adding normally distributed noise to random points on the curve.
The curve, a contaminated point cloud and results of application of both the kernel PCA related basic approach from Section~\ref{sec:the_basic_technique} and the new proposed technique are illustrated on Figure~\ref{fig:toy_example}.
In this Figure the kernel PCA based approach is virtually equivalent~(see Appendix \ref{appdx:connection_to_PCA} for the details) to minimizing the $l_2$-norm of the algebraic distances with Euclidean orthonormality constraints, while the proposed technique utilizes $l_1$-norm minimization of the $2$-distances with the Hilbert-Schmidt orthonormality constraints.

\subsection{Outlier Detection for Face Recognition}
\label{sec:exp_ood_and_robust}

We consider an image classification pipeline consisting of a face detection and alignment algorithm (MTCNN, \textcite{xiang2017joint}), deep feature extractor (ArcFace, \textcite{deng2019arcface}) and cosine similarity based classifier. The feature extractor used was trained on the variation of the MS1M dataset \cite{guo2016msceleb1m}.\footnote{We used the MS1M-ArcFace dataset from \url{https://github.com/deepinsight/insightface/wiki/Dataset-Zoo}. In~the~sequel, for outliers we use the Anime-Faces dataset from \url{https://github.com/bchao1/Anime-Face-Dataset}.}
In this setting, we fit a quadric intersection manifold to $\approx 6\cdot 10^6$ feature space embeddings of photos from the same dataset. The embeddings lie in the $512$-dimensional space. Details of the test datasets construction and respective licences are discussed in Appendix~\ref{appdx:experiments}. We then measure the performance of the quadric intersection based outlier detector and the performance of the cosine similarity improved by penalizing the outliers detected by it as by Section~\ref{subsec:similarity_robust}.
The quadric based approach is compared to various geometry-based outlier detectors such as principal component analysis (PCA), kernel principal component analysis for novelty detection (KPCA-ND, \textcite{hoffmann2007kernel}) with RBF kernel and kernel one-class support vector machine (OCSVM, \textcite{scholkopf1999support}) with degree $3$ polynomial kernel.\footnote{In our setting, OCSVM is also equivalent to support vector data description \cite{lampert2009kernel}.}
In~all~these approaches we normalize the embeddings as a preprocessing step.
Finally, we compare our technique to the approach based on the embedding norm (NORM), that is motivated by recent work of \textcite{yu2020out} that shows that in the face recognition domain, the norm of an embedding might carry some information on the image outlierness.

Motivated by a simple ablation study (see details in Appendix~\ref{appdx:ablation}), we fit the intersection of $100$ quadrics~(we refer to this approach as Q-FULL) and a $170$-dimensional PCA plane to data.
Additionally, to showcase the advantages of the proposed technique over the basic idea from Section~\ref{sec:the_basic_technique}, we fit the intersection of $100$ quadrics by applying SGD to solve the kernel PCA related optimization problem~\eqref{eqn:alg_dist_optimization}, we refer to this approach as Q-BASE.
Since the na\"ive kernel PCA scales cubically with respect to data size, to use the RBF-based KPCA-ND we need to resort to approximations. Specifically, we utilize $300$ random Fourier features \cite{rahimi2007random} to approximate the implicit feature map.
OCSVM does not scale favorably with data size as well, thus we train it on a subsample of size $8\cdot 10^{4}$ of the full training data.
To examine our method in the low-data regime we also consider an intersection of $30$ quadrics fit to the same subsample of size $8\cdot 10^{4}$ of the data, we refer to this as Q-SUB.
\begin{table*}[t]
\caption{
The AUC-ROC scores for different feature space outlier detectors described in the main text.
For the larger datasets (all except CPLFW and CALFW) the order of standard deviation was estimated from $10$ random subsamplings.
For OCSVM the standard deviation is of order $10^{-2}$, while for all other methods it does not exceed a number of order $10^{-3}$.
Here (a), (b), (c), (d), (e) correspond respectively to \textcite{kemelmacher2016megaface}, \textcite{cao2018vggface2}, \textcite{karras2019style}, \textcite{zheng2018cross}, \textcite{zheng2017cross}. (b) has been taken from \url{https://github.com/deepinsight/insightface/wiki/Dataset-Zoo}.
}
\label{table:AUC}
\centering
\begin{tabular}{ r c c c c c c c } 
Dataset & \tableoneorder{Q-FULL}{Q-BASE}{Q-SUB}{PCA}{KPCA-ND}{OCSVM}{NORM}\\
\midrule
MS1M-ArcFace & \tableoneorder{\textbf{0.97}}{0.95}{0.82}{0.89}{0.66}{0.71}{0.75}\\
MegaFace (a) & \tableoneorder{\textbf{0.89}}{0.87}{0.76}{0.81}{0.73}{0.76}{0.74}\\
\!\!\!VGGFace2 (b) & \tableoneorder{\textbf{0.96}}{0.94}{0.81}{0.88}{0.70}{0.83}{0.75}\\
FFHQ     (c) & \tableoneorder{\textbf{0.93}}{0.92}{0.82}{0.90}{0.71}{0.85}{0.72}\\
CPLFW    (d) & \tableoneorder{\textbf{0.93}}{0.91}{0.73}{0.81}{0.67}{0.82}{0.75}\\
CALFW    (e) & \tableoneorder{\textbf{0.98}}{0.95}{0.79}{0.88}{0.71}{0.84}{0.70}\\
\bottomrule
\end{tabular}
\end{table*}

Fitting an intersection of $100$ quadrics took us $72$ hours on a pair of Quadro RTX 8000 GPUs, using an unoptimized implementation, for both methods Q-BASE and Q-FULL.
The resulting (weights of the) quadric intersection models occupy around $50$MB each.

\begin{table*}[b]
\caption{
Effect of the various modifications of similarity on the identification rate. Methods names are as in Table \ref{table:AUC}. False positive rate is fixed to $10^{-5}$. The corresponding threshold hyperparameters are given in Appendix \ref{appdx:ir_hyperparameters}.
}
\label{table:IR}
\centering
\begin{tabular}{ r c c c c c c c c }
Metric & Initial & \tabletwoorder{Q-FULL}{Q-SUB}{PCA}{KPCA-ND}{Q-BASE}{OCSVM}{NORM}\\
\midrule
Full IR & 0.61 & \tabletwoorder{\textbf{0.67}}{0.61}{0.65}{0.59}{0.63}{0.64}{\textbf{0.67}}\\
IR      & 0.66 & \tabletwoorder{0.74}{0.66}{0.72}{0.64}{0.69}{0.70}{\textbf{0.75}}\\
\bottomrule
\end{tabular}
\end{table*}

\paragraph{Outlier Detection} 

We construct a contaminated dataset by mixing the in-distribution photos from one of the special face recognition datasets with the out-of-distribution photos in the approximate ratio of $99$ to $1$.
The out-of-distribution photos contain $235$ manually picked photos from the CPLFW dataset \cite{zheng2018cross} where a face cannot be uniquely recognized by a human (e.g. photos of people in hockey helmets, photos with multiple faces) and $235$ images from the Anime-Faces dataset\footnotemark[7] aligned by the MTCNN.
See the resulting AUC-ROC values for different detectors in Table~\ref{table:AUC}.

\paragraph{Similarity Robustification}
Here we evaluate performance of the similarity-based classifier with robustified similarity function. All robustification methods are based upon the equation \eqref{eqn:robustification_truncation}: the outlier score from the corresponding detector is used for thresholding the similarity between given embeddings. For the test scenario we consider the previously-mentioned Cross-Posed Labeled Faces in the Wild (CPLFW) dataset which is considered particularly challenging due to the presence of the pictures that cannot be recognized even by a human (which we consider as outliers). Performance is measured in terms of the identification rate, which can be understood as the true positive rate of the similarity-based classifier solving an identification problem in the presence~(Full~IR) or in the absence~(IR) of distractors taken from the MegaFace dataset, see details in Appendix \ref{appdx:IR}.

Both CPLFW and MegaFace are split in two halves.
The first half is used for choosing the threshold hyperparameter $t$ and the other half is used to measure performance (in terms of the identification rate).
The Full IR and IR corresponding to each of the robustification methods is presented in Table \ref{table:IR}.
Q-FULL, Q-BASE and NORM perform similarly in this experiment, outperforming other methods.

\subsection{Discussion and Method Limitations}
Quadric based techniques behave favorably in both the outlier detection and similarity metric robustification problems, improving on the classic baselines.
The norm based approach turns out to be a stronger competitor, which is not surprising given that it is specialized to the setting at hand.
Our approach, which is generic, matches its performance in similarity metric robustification, and improves upon its performance in outlier detection.
The performance of the Q-SUB approach reveals the limitation of our approach: in the low-data regime quadrics-based model is outperformed by the classical geometry-based outlier detection approaches.
This is to be expected though, as the proposed technique is designed for use within the big data domain.

\section{CONCLUSION}
We describe a manifold learning technique based on fitting an intersection of quadrics to a point cloud.
To make the problem of fitting a quadric intersection to data tractable, we start from the simplest possible approximate formulation that turns out to be deeply related to the polynomial~kernel~PCA.
Analysing its downsides, we proceed to introduce a number of improvements to the formulation that promote robustness and equivariance.
The resulting optimization problem is tractable in moderately high dimension, such as the feature space of a deep representation learning model, is amenable to minibatch training, and thus scales well with respect to point cloud size.
The learned quadric intersection can be used to define an outlier score and to improve a given similarity metric.
We demonstrate and benchmark the proposed approach empirically on an open set image classification~task.
 
\subsection{Societal impact}\label{sec:broader_impact}

The paper is mainly theoretical, presenting a new manifold learning technique suitable for modern application settings, mainly outlier detection and similarity metric improvement at the feature space level of deep representation learning models. The pipelines based on these models are widespread in computer vision and natural language processing where the scalability of the proposed technique allows it to be used as a drop in solution in a wide range of industrial applications.

As the main application of the approach is in the domain of anomaly detection, technique may be used to increase reliability of existing pipelines.
Our experiments have shown that our technique may be used to improve the identification rate within a facial recognition framework. Examples of the negative societal impact of misuse of such frameworks are widely known.
However, we stress that the technique's advantages are simplicity and generality, and our choice of the experimental setting should not be regarded as determinative.

\section*{Acknowledgments}

FP was supported within the framework of the Basic Research Program at HSE University.
SI and AZ were supported by the Ministry of Science and Higher Education of the Russian Federation, agreement N\textsuperscript{\underline{o}} 075-15-2019-1619.
 VB was supported by the Ministry of Science and Higher Education of the Russian Federation, agreement N\textsuperscript{\underline{o}} 075-15-2019-1620.

\printbibliography


\onecolumn
\appendix
\clearpage
\newpage
\section{Theory}\label{appdx:theory}
\subsection{Connection to polynomial kernel PCA}\label{appdx:connection_to_PCA}
Consider a feature map $\v{\varphi} : \R^d \!\to\! \R^{D}$ that is given by
\[ \label{eqn:feature_transform}
\v{\varphi}(p_1,\dots,p_d)
=
(\underbrace{\dots,p_ip_j,\dots}_{\text{pairwise products}},\underbrace{p_1,\dots,p_d\vphantom{p_j}}_{\text{coordinates}},\underbrace{ 1 \vphantom{p_j}}_{\text{constant}})^\top
.
\]
For a quadratic polynomial $f$ we have \[ \label{eq:f(p)}
f(\v{p})
=
\langle \v{\varphi}(\v{p}), \v{v}(f) \rangle
,
\]
where $\v{v}(f)$ is the coefficient vector (see Section \ref{sec:the_basic_technique}). 

Let $f_1,\dots,f_m$ be quadratic polynomials in $\R^d$ and $\v{v}(f_1),\dots,\v{v}(f_m)\in \R^D$ be their coefficient vectors that we assume to be orthonormal.
We denote by $\c{V}_m \subseteq  \R^D$ the vector space spanned by these vectors and by $\c{V}_m^\bot$ its orthogonal completion.  A simple computation (see Appendix~\ref{appdx:kpca} for details) yields
\[ \label{eq:sum_of_squares_of_values}
\sum_{j=1}^n \sum_{k=1}^m \abs{f_k(\v{p}_j)}^2
=\sum_{j=1}^n \distgeom(\v{\varphi}(\v{p}_j), \c{V}_m^\bot)^2.
\]
It follows that the optimization problem is equivalent to minimization of $\sum_{j=1}^n \distgeom(\v{\varphi}(\v{p}_j), \c{V}_m^\bot)^2$, the same problem that PCA solves.
This means that the technique presented above is equivalent to applying PCA in a feature space defined by~\eqref{eqn:feature_transform}, when viewed from a different angle. Note that here we do not assume that the point cloud $\v{\varphi}(\v{p}_1),\dots, \v{\varphi}(\v{p}_n)$ is centered, so the optimization problem \eqref{eqn:alg_dist_optimization} is equivalent to the non-centered version of PCA.

A slight modification of the feature map \eqref{eqn:feature_transform} that is given by
\begin{equation}
\tilde{\v{\varphi}}(p_1,\dots,p_d) =
(\dots,p_ip_j,\dots,\sqrt{2}p_1,\dots,\sqrt{2}p_d, 1)^\top    
\end{equation}
corresponds to the polynomial kernel $k(\v{x}, \v{y}) = \del{\innerprod{\v{x}}{\v{y}} + 1}^2$~of degree $2$ in the sense that $k(\v{x}, \v{y}) = \innerprod{\tilde{\v{\varphi}}(\v{x})}{\tilde{\v{\varphi}}(\v{y})}$.
This shows that the optimization problem is closely connected with kernel PCA with a polynomial kernel.\footnote{The feature map \eqref{eqn:feature_transform} is also similar to the Veronese map of degree $2$ used in Generalized PCA \cite[\S 3.1]{vidal2005generalized}. The difference is we consider not only quadratics, but all monomials of degree $\leq 2$.}

This suggests an alternative way of solving the optimization problem~\eqref{eqn:alg_dist_optimization}---by computing the SVD of a $D \x D$ matrix, although it is usually impractical since $D$ depends quadratically on $d$ and the computational complexity of SVD is of order $O(D^3) = O(d^6)$.

\subsection{Additional details on the connection to polynomial kernel PCA} \label{appdx:kpca}

For each $\v{u}\in \R^D$ we denote by ${\sf pr}(\v{u})$ its orthogonal projection onto $\c{V}_m$.
It is defined by the formula
\[
{\sf pr}(\v{u})
=
\sum_{k=1}^m \innerprod{\v{u}}{\v{v}(f_k)} \cdot \v{v}(f_k)
.
\]
Therefore the following holds for the distance $\distgeom(\v{u}, \c{V}_m^\bot)$ form $\v{u}$ to $\c{V}_m^\bot$:
\[\label{eq:norm_of_projection}
\distgeom(\v{u}, \c{V}_m^\bot)^2
=
\norm{{\sf pr}(\v{u})}^2 
= 
\sum_{k=1}^m \langle \v{u},\v{v}(f_k)\rangle^2
.
\]
Using the formulas \eqref{eq:norm_of_projection} and \eqref{eq:f(p)} we obtain
\[
\sum_{j=1}^n \sum_{k=1}^m \abs{f_k(\v{p}_j)}^2
=
\sum_{j=1}^n \sum_{k=1}^m \innerprod{\v{\varphi}(\v{p}_j)}{\v{v}(f_k)}^2
=
\sum_{j=1}^n \norm{{\sf pr}(\v{\varphi}(\v{p}_j)) }^2
=
\sum_{j=1}^n \distgeom(\v{\varphi}(\v{p}_j),\c{V}_m^\bot)^2
,
\]
which proves equation~\eqref{eq:sum_of_squares_of_values}.

\subsection{Non-equivariance of the optimization problem \eqref{eqn:alg_dist_optimization}}\label{appdx:non-equivariance}

\begin{figure}[H]
\centering
\begin{subfigure}{0.3\textwidth}
\includegraphics[width=\textwidth]{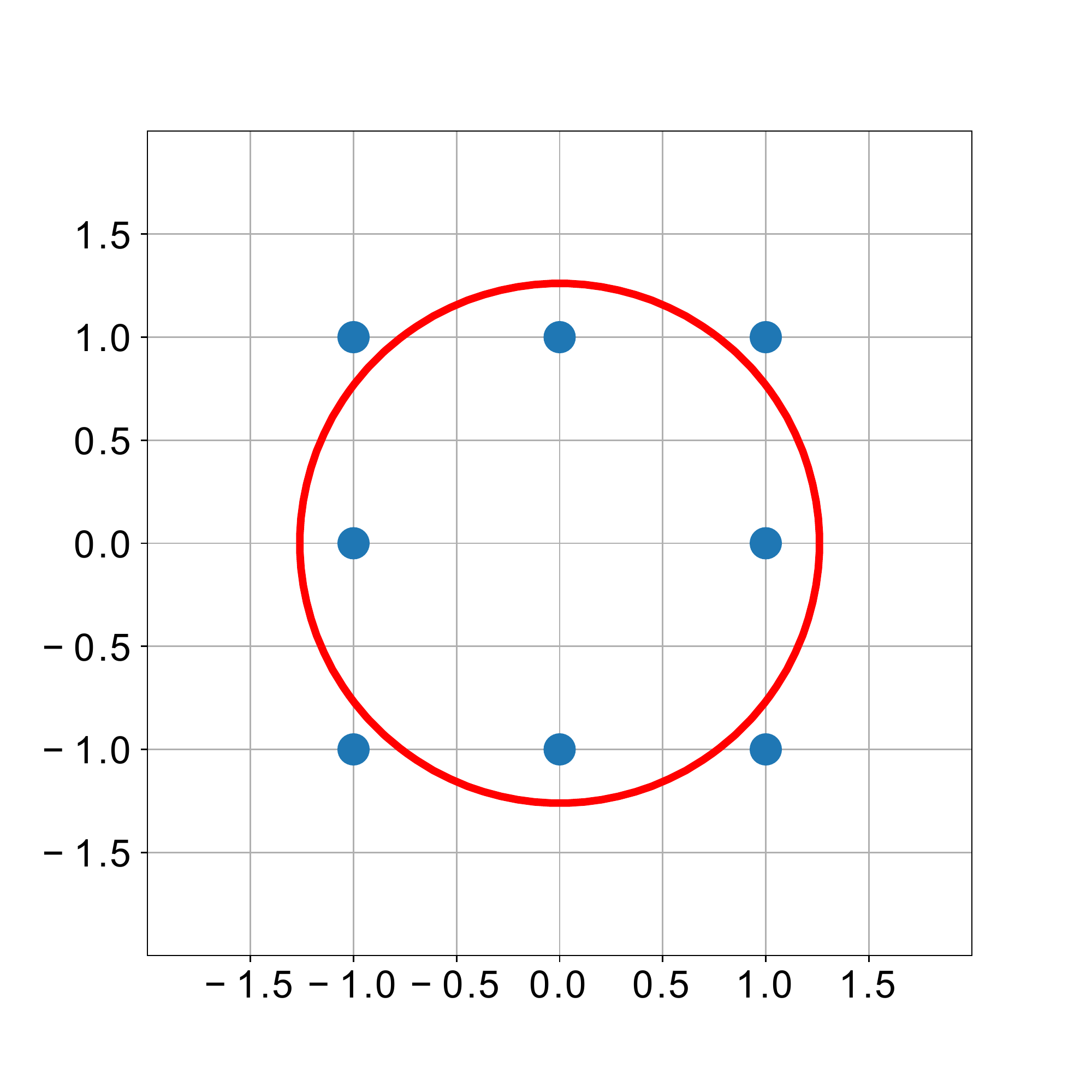}
\caption{The original point cloud.}
\end{subfigure}
\hspace{2cm}
\begin{subfigure}{0.3\textwidth}
\includegraphics[width=\textwidth]{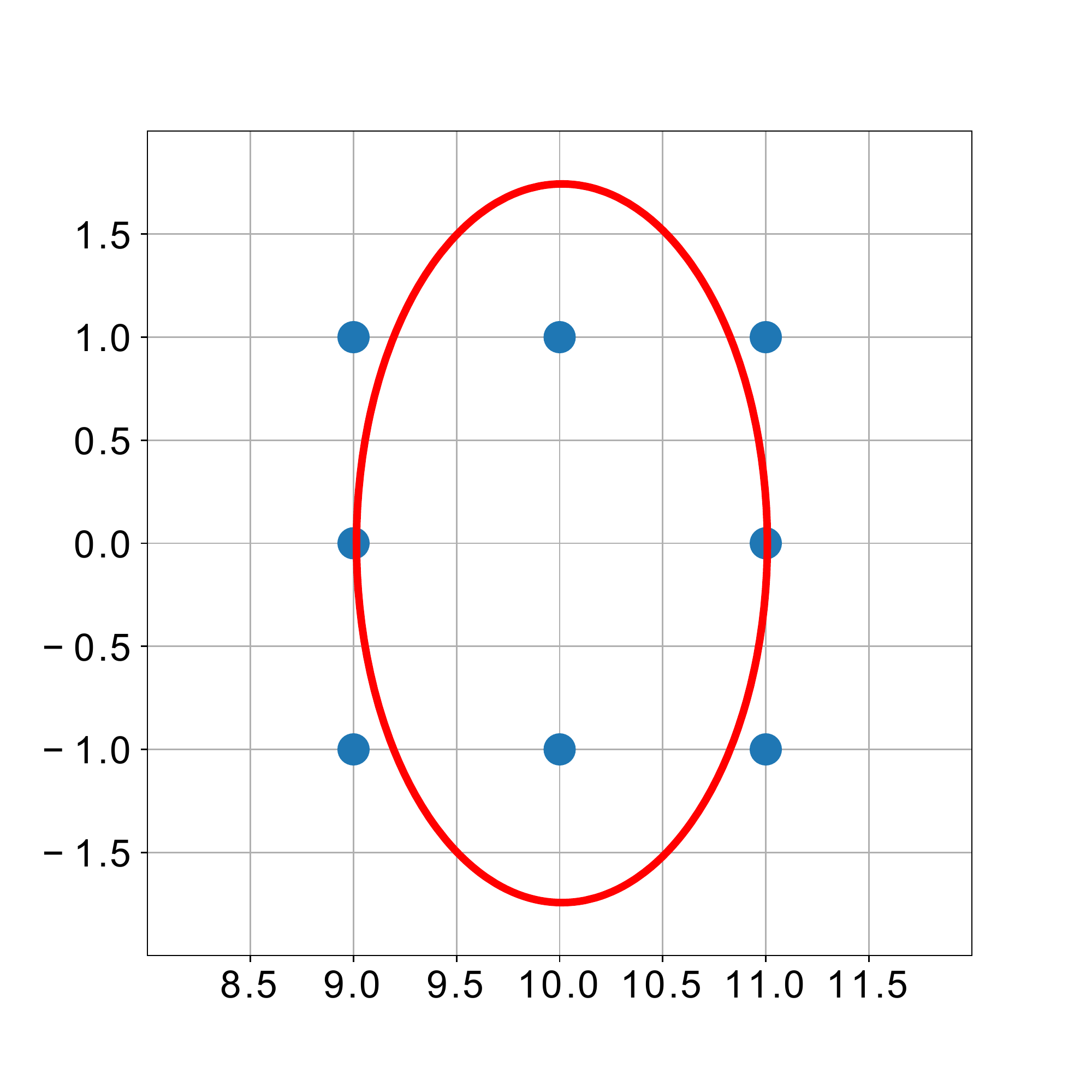}
\caption{The shifted point cloud.}
\end{subfigure}
\caption{A point cloud and its shifted version (blue) and the respective quadrics (red) that solve the optimization problem~\eqref{eqn:alg_dist_optimization}.}
\label{fig:non_equivariance_example}
\end{figure}

On the pictures below we show an example of two point clouds on the plane and the corresponding quadrics that solve the optimization problem~\eqref{eqn:alg_dist_optimization} exactly.
The point cloud on Figure~\ref{fig:non_equivariance_example} (b) can be obtained by shifting the point cloud on Figure~\ref{fig:non_equivariance_example} (a) by vector $(10,0)$.
However, the corresponding quadrics are not the shifts of each other.

The picture corresponding to the optimization problem from Section~\ref{sec:improved_technique} should be similar to Figure~\ref{fig:non_equivariance_example}~(a) because of the equivariance: the solution of the translated problem is the translated solution of the original problem.

\subsection{Approximation distance of order $k$} \label{appendix_distances}

Let $f:\R^d\to \R$ be a polynomial.
For a multi-index $I=(i_1,\dots,i_d)$ we denote by $C_I$ the coefficient of the Taylor polynomial of $f$ in the point $p$ corresponding to the monomial $x_1^{i_1}\dots x_d^{i_d}:$
\[
C_I= \frac{1}{I!} \frac{\partial^{|I|} f}{ \partial x_1^{i_1} \dots \partial x_d^{i_d} }(p),
\]
where $|I|=i_1+\dots+i_d$ and $I! = i_1!i_2!\dots i_d!$.

For each integer $l\geq 1$ we set
\[\label{eq_c_l}
c_l
=
-\left({\sum}_{|I|=l} C_I^2/b(I)\right)^{1/2}
,
\]
where $b(I)={\abs{I}!}/{ I! }$ is the multinomial coefficient and $c_0=|f(\v{p})|$.
Note that $c_0\geq 0$ and $c_l\leq 0$ for $l\geq 1$.
With this, the polynomial $\sum_{l=0}^k c_lt^l$ has a unique non-negative root.
We denote this root by $\distk{k}(\v{p},Z(f))$ and, following \textcite{taubin1993improved}, call it
{\it the approximation distance of order $k$}.
Note that $\distk{k}(\v{p},Z(f))$ depends on $\v{p}$ and $f$, not only on $\v{p}$ and the zero set $Z(f)$, despite the notation.

If $f(x)=\v{x}^\top \m{A} \v{x} +{\bf b}x+{\bf c}$ is a quadratic polynomial, then it is easy to check that 
$c_2=\|\m{A} \|_{HS}.$ Using this and the formula for the roots of the general quadratic equation we obtain the formula
\[
\disttwo(\v{p},Z(f))
=
\frac{\sqrt{h^2 + \abs{f(\v{p})} \norm{f}_{HS}} - h}{\norm{f}_{HS}},
\]
where $h = \norm{\grad f(\v{p})}/2.$ In particular, if $\norm{f}_{HS}=1,$ we have a particularly simple expression
\[
\disttwo(\v{p},Z(f))
=
\sqrt{h^2 + \abs{f(\v{p})}} - h
.\]

\subsection{Equivariance of the approximation distance order $k$}
\label{appdx:equivariance}
The coefficients of the polynomial $c_0+c_1t+\dots+c_kt^k$ from the definition of $\distk{k}(\v{p},Z(f))$ depend only on the coefficients of the Taylor polynomial of degree $k$ of the map $f$ at point $\v{p}$.
We denote this polynomial by $T_{k,f,\v{p}}$.
In order to prove the equivariance of the $k$-distance, i.e.
$
\distk{k}(\theta(\v{p}), Z(f\circ \theta) ) =
\distk{k}(\v{p},Z(f))
,
$
it is enough to show the equivariance of the Taylor polynomial, in the sense that
\[
T_{k,\v{p},f}\circ \theta = T_{k, f\circ \theta,\theta(\v{p})}.
\]

The defining property of the Taylor polynomial is the following: $T_{k,f,\v{p}}$ is the only polynomial of degree $\leq k$ such that
\[\label{eqn:taylor}
\norm{T_{k,f,\v{p}}(\v{x}) - f(\v{x})}
=
o(\norm{\v{x}-\v{p}}^k)
\hspace{1cm} \t{as} \   \v{x} \to \v{p}
.
\]
Any isometry $\theta:\R^d\to \R^d$ is of form $\theta(\v{x})=\m{Q}\cdot \v{x} + \v{v},$ where $\m{Q}$ is an orthogonal matrix and $\v{v}$ is some vector.
Hence~$T_{k,f,\v{p}}\circ \theta$ is also a polynomial of degree $\leq k$.
The equation \eqref{eqn:taylor} then implies
\[
\norm{(T_{k,f,\v{p}}\circ \theta) (\v{x}) - (f\circ \theta) (\v{x})} =
o(\norm{\theta(\v{x})-\theta(\v{p})}^k)
=
o(\norm{\v{x}-\v{p}}^k)
\hspace{1cm} \t{as} \   \v{x}\to \v{p}
.
\]
Therefore $T_{k,\v{p},f}\circ \theta = T_{k, f\circ \theta,\theta(\v{p})}$.

\subsection{Equivariance of the Hilbert--Schmidt inner product} \label{appdx:equivariance_HS}

The Hilbert--Schmidt inner product of matrices can be written as
\[ 
\langle \m{A},\m{B} \rangle_{HS} = {\sf tr}(\m{A}^\top \cdot \m{B})= {\sf tr}(\m{B}^\top \cdot \m{A}),\]
where ${\sf tr}(\cdot)$ denotes the trace of a matrix. 
It follows that for any orthogonal matrix $\m{Q}$ the following holds
\[
\label{eqn:eqiv_HS}
\langle \m{A},\m{B} \rangle_{HS} = \langle \m{A}\m{Q},\m{B}\m{Q} \rangle_{HS} = \langle \m{Q}\m{A},\m{Q}\m{B} \rangle_{HS}.
\]

Recall that any isometry $\theta:\R^d\to \R^d$ is of form $\theta(\v{x})=\m{Q}\cdot \v{x} + \v{v},$ where $\m{Q}$ is an orthogonal matrix and $\v{v}$ is a vector.
If~$f$ is a quadratic polynomial with the corresponding matrix $\m{A}$ (in the sense of equation~\eqref{eqn:polynomial_matrix_form}), then $f\circ \theta $ is a quadratic polynomial whose corresponding matrix equals to
$\m{Q}^\top \m{A} \m{Q}.$ Therefore the equation \eqref{eqn:eqiv_HS} implies
\[\langle f \circ \theta ,g\circ \theta \rangle_{HS} = \langle \m{Q}^\top \m{A}\m{Q}, \m{Q}^\top \m{B}\m{Q} \rangle_{HS} = \langle \m{A},\m{B} \rangle_{HS}= \langle f,g \rangle_{HS}.\]

\section{Additional experimental details}\label{appdx:experiments}

\subsection{Datasets licenses}\label{appdx:licenses}
\begin{enumerate}
    \item MS1M-ArcFace was derived from MS1M dataset by InsightFace project, the license of the project applies: \url{https://github.com/deepinsight/insightface}. 
    \item Images of MegaFace are licensed under Creative Commons License, details of dataset license are given in \url{http://megaface.cs.washington.edu/dataset/download.html}.
    \item VGGFace2 is licensed under Creative Commons Attribution 4.0 International license, details are given in \url{https://web.archive.org/web/20171113123726/http://www.robots.ox.ac.uk/\~vgg/data/vgg\_face2/licence.txt}.
    \item FFHQ is licensed under Creative Commons BY-NC-SA 4.0 license by NVIDIA Corporation, details are given in \url{https://github.com/NVlabs/ffhq-dataset/blob/master/LICENSE.txt}.
    \item The licences for CPLFW, CALFW and Anime-Faces datasets are unknown to the authors. The official dataset websites \url{http://www.whdeng.cn/cplfw/index.html}, \url{http://whdeng.cn/CALFW/index.html} and \url{https://github.com/bchao1/Anime-Face-Dataset} do not provide license information.
\end{enumerate}

\subsection{Datasets construction}
The embeddings are constructed by the pretrained ArcFace model LResNet100E-IR, ArcFace@ms1m-refine-v2\footnote{https://github.com/deepinsight/insightface/wiki/Model-Zoo}.
For MegaFace, FFHQ, CALFW and CPLFW datasets additional alignment was performed by by MTCNN.
For CPLFW, due to the complex nature of the dataset, the photos where MTCNN failed to detect a face were preserved in the dataset, with no additional preprocessing applied apart from resizing.
Aligned versions of MS1M-ArcFace and VGGFace2 datasets were taken from \url{https://github.com/deepinsight/insightface/wiki/Dataset-Zoo}.

\subsection{Identification rate}\label{appdx:IR}

Here we describe the procedure for evaluating the performance of the similarity function $s$ on the identification problem on the set $X$ of face images with the additional set $Y$ of distractor images.
It is based on the metric called \emph{identification rate} and is widely used in the face recognition domain \cite{liao2014benchmark}.

First, define a family of classifiers $C_a$, parameterized by $a\in \R$ as follows: $C_a$ declares a pair $p =(x,y)$ positive (same class), if $s(x,y)\geq a$.
The corresponding false positive rate and true positive rate are denoted by ${\sf fpr}(a)$ and ${\sf tpr}(a)$ respectively.
For a fixed false positive rate $f\in [0,1]$ we define the similarity threshold ${\sf sth(f)}$ by
\[
{\sf sth(f)}
=
\sup\Set*{a \in \R}{{\sf fpr}(a) \leq f}.
\]
The \emph{identification rate} is defined to be ${\sf tpr}({\sf sth}(f))$.

Consider the set
\[
Q
=
\Set*{(x,y)\in X\times X}{\forall d\in Y\ \  s(x,y) > \max (s(x,d), s(y,d))}
,
\]
which contains pairs \emph{that cannot be distracted} by elements of $Y$ and consider the binary classifier that declares a pair $(x,y)$ positive (same class) if it is positive with respect to the classifier $C_{\sf sth(f)}$ and it cannot be distracted.
The corresponding ${\sf tpr}$ is called the \emph{full identification rate}.
We note that another known terminology for IR and Full IR are \emph{verification rate} and \emph{identification rate} respectively.
This terminology comes from the corresponding problems in face recognition: 
\begin{enumerate}
    \item The verification problem is to determine from a pair of images whether they are photos of the same person. This problem is typically solved by introducing a similarity measure between images or their embeddings. In a given benchmark like CPLFW, the verification rate is calculated using the pairs suggested by the dataset creators. However, in our experiments we use all possible pairs in CPLFW.
    \item The identification problem is to determine the identity of a person on a given image. This problem is typically posed in the setting with distractors.
\end{enumerate}

In our experiments the subset of first $8 \cdot 10^5$ embeddings of the MegaFace dataset is used as the distractor set $Y$.

\subsection{Ablation study}\label{appdx:ablation}

The results of the simple ablation study used to identify the best number of quadrics in intersection and the optimal number of principal components in the setting of Section~\ref{sec:exp_ood_and_robust} are presented here.
AUC-ROC and IR scores are given in Table~\ref{tab:ood_ablation} and Table~\ref{tab:ir_ablation} respectively.
The preliminary selection of numbers of principal components were made by studying the eigenvalue decay.

\begin{table}[h!]
\caption{AUC-ROC scores for the quadrics based outlier detector depending on the number of quadrics in the intersection and AUC-ROC scores for the PCA based outlier detector depending on the number of principal components. Standard deviations do not exceed $10^{-3}$.}
\label{tab:ood_ablation}
\centering
\begin{tabular}{rccccccc}
\toprule
& \multicolumn{3}{c}{Quadrics} & \multicolumn{3}{c}{PCA} \\ \cmidrule(r){2-4} \cmidrule(r){5-7}
Dataset & 50 quadrics & 100 quadrics & 200 quadrics & 130-dim &                  170-dim & 200-dim \\
\midrule
MS1M-ArcFace & 0.97  & 0.97  & 0.97  & 0.88   & 0.89    & 0.87   \\
MegaFace & 0.88  & 0.89  & 0.89  & 0.79   & 0.81    & 0.81   \\
VGGFace2 & 0.95  & 0.96  & 0.96  & 0.86   & 0.88    & 0.86   \\
FFHQ & 0.92  & 0.93  & 0.93  & 0.89   & 0.90    & 0.89   \\
CPLFW & 0.92 & 0.93 & 0.93  & 0.80 & 0.82 & 0.80 \\
CALFW & 0.97 & 0.98 & 0.97  & 0.88 & 0.88 & 0.85 \\
\bottomrule
\end{tabular}
\end{table}

\begin{table}[h!]
\caption{IR score for the face identification problem with quadric-based and PCA-based robustification depending (respectively) on the number of quadrics in the intersection or on the number of principal components.}
\label{tab:ir_ablation}
\centering
\begin{tabular}{lccccccc}
\toprule
& \multicolumn{3}{c}{Quadrics} & \multicolumn{3}{c}{PCA} \\ \cmidrule(r){2-4} \cmidrule(r){5-7}
Metric & 50 quadrics & 100 quadrics & 200 quadrics & 130-dim &                  170-dim & 200-dim \\
\midrule
Full IR & 0.632 & 0.635 & 0.635 & 0.620 &  0.626 & 0.710 \\
IR & 0.737 & 0.741 & 0.742 & 0.611 & 0.720 & 0.700\\
\bottomrule
\end{tabular}
\end{table}
Another way to guess the optimal number of quadrics is to study the eigenvalues of PCA by means of Wold invariant \cite{wold1978cross} or plotting and analyzing $-\log \lambda_n/\lambda_1$.
If the estimated dimension of the linear manifold is $d-k$, then $k$ may serve as the number of quadrics or as its initialization for further tuning. The problem-specific limitation on the model size that often exists in practice may also give an upper bound on the number of quadrics.
\subsection{Methods implementation details and hyperparameters}\label{appdx:ir_hyperparameters}

Quadrics intersection fitting was implemented in \texttt{torch} framework, the corresponding model and training routines are included in the repository. PCA and one-class SVM methods implementations from \texttt{sklearn} library were used. For kernel PCA based novelty detector the implementation from \url{https://github.com/Nmerrillvt/kPCA} was used together with random Fourier features implementation from \url{https://github.com/tiskw/random-fourier-features}.

The constrains in the optimization problem \eqref{eq:our_optimization_problem} require optimization over the set of orthonormal frames relation.
Because of this, stochastic gradient descent over the Stiefel manifold could be a natural choice for solving the constrained optimization problem.
However, in the preliminary experiments we observed that the simple soft regularization (as in Equation~\eqref{eq:concrete_loss}) with $\lambda = 1$, is more effective and efficient compared to the manifold optimization.
The latter does not requre tuning $\lambda$ but may introduce other algorithm-specific hyperparameters.
In our experiments the error term $\norm[0]{\tilde V(\v{F})^T \tilde V(\v{F}) \!-\! I}^2_{HS}$ was of order $10^{-5}$, which was small enough for our purposes.

The typical values of the threshold hyperparameter $t$ in the similarity robustification experiment are given in Table~\ref{tab:hyperparams}.
They were determined in a series of $10$ experiments.
In each experiment, both CPLFW and MegaFace were randomly split in two halves.
First half was used for determining threshold $t$ by means of a grid search and the second one was reserved for evaluating the identification rate.

Robustification based on Q-SUB and KPCA-ND methods leads to IR deterioration for all possible values of the threshold hyperparameter. For other methods, optimal values of threshold correspond to marking around one percent of data as outliers. Because of this, to make comparisons fair, we choose thresholds for Q-SUB and KPCA-ND so as to get one percent of outliers, matching the behavior of other methods.

\begin{table}[b]
\caption{Threshold hyperparameter values for various similarity robustification methods.}
\label{tab:hyperparams}
\centering
\begin{tabular}{ r c c c c c c c }
 & \tabletwoorder{Q-FULL}{Q-SUB}{PCA}{KPCA-ND}{Q-BASE}{OCSVM}{NORM}\\
\midrule
Threshold  & \tabletwoorder{0.033}{0.452}{0.529}{31.47}{0.163}{$2.9\cdot 10^{-9}$}{14.55}\\
\bottomrule
\end{tabular}
\end{table}

\end{document}